%% file: main.tex
\documentclass{article}
\usepackage[dvipsnames]{xcolor}         
\definecolor{linkColor}{rgb}{0.18,0.39,0.62}
\usepackage[colorlinks=true,linkcolor=linkColor,citecolor=linkColor,filecolor=linkColor,urlcolor=linkColor]{hyperref}       

\usepackage[final]{neurips_2023}

\usepackage[utf8]{inputenc} 
\usepackage[T1]{fontenc}    
\usepackage{url}            
\usepackage{booktabs}       
\usepackage{amsfonts}       
\usepackage{nicefrac}       
\usepackage{microtype}      
\usepackage{xcolor}         

\usepackage{subcaption}
\input{math_commands.tex}

\usepackage{hyperref}
\usepackage{url}
\usepackage{graphicx}
\usepackage{arydshln}
\usepackage{booktabs}
\usepackage{multirow}
\usepackage{caption}
\usepackage{makecell}
\usepackage{float}

\newcommand\our{\textsc{Promptist}}

\title{Optimizing Prompts for Text-to-Image Generation}

\author{\\
\textbf{Yaru Hao\thanks{~Equal contribution.},~~\ Zewen Chi\footnotemark[1],~~\ Li Dong,~~Furu Wei} \\
Microsoft Research \\
\url{https://github.com/microsoft/LMOps} \\}

\begin{document}

\maketitle

\begin{abstract}
Well-designed prompts can guide text-to-image models to generate amazing images. However, the performant prompts are often model-specific and misaligned with user input. Instead of laborious human engineering, we propose prompt adaptation, a general framework that automatically adapts original user input to model-preferred prompts. Specifically, we first perform supervised fine-tuning with a pretrained language model on a small collection of manually engineered prompts. Then we use reinforcement learning to explore better prompts. We define a reward function that encourages the policy to generate more aesthetically pleasing images while preserving the original user intentions. Experimental results on Stable Diffusion show that our method outperforms manual prompt engineering in terms of both automatic metrics and human preference ratings. Moreover, reinforcement learning further boosts performance, especially on out-of-domain prompts. 
The pretrained checkpoints are available at \url{https://aka.ms/promptist}. The demo can be found at \url{https://aka.ms/promptist-demo}.
\end{abstract}

\section{Introduction}

Generative foundation models can be prompted to follow user instructions, including language models~\citep{gpt3, palm, mtnlg}, and text-to-image models~\citep{dalle, dalle2, imagegen, sdiffusion}.
It has been recognized that prompt design plays an essential role in the generation quality. We need to adjust the prompt to make the model better understand our intentions and produce higher-quality results~\citep{promptprogramming, lmengineer}.
The problem is severe in text-to-image models because the capacity of their text encoders, such as CLIP text encoder~\citep{clip} in Stable Diffusion~\citep{sdiffusion}, is relatively small.
Empirical observations also confirm that common user input is often insufficient to produce aesthetically pleasing images with current models.

Prior efforts implement manual prompt engineering towards specific text-to-image models~\citep{promptguideline, promptmodifier, guidebook}, typically adding some modifiers to the original input.
However, it is laborious and sometimes infeasible to conduct manual prompt engineering.
Besides, the manually engineered prompts often cannot be transferred between various model versions. 
Therefore, it is necessary to find a systematic way to automatically align user intentions and various model-preferred prompts.

\begin{figure*}[t]
\centering
\includegraphics[width=0.99\textwidth]{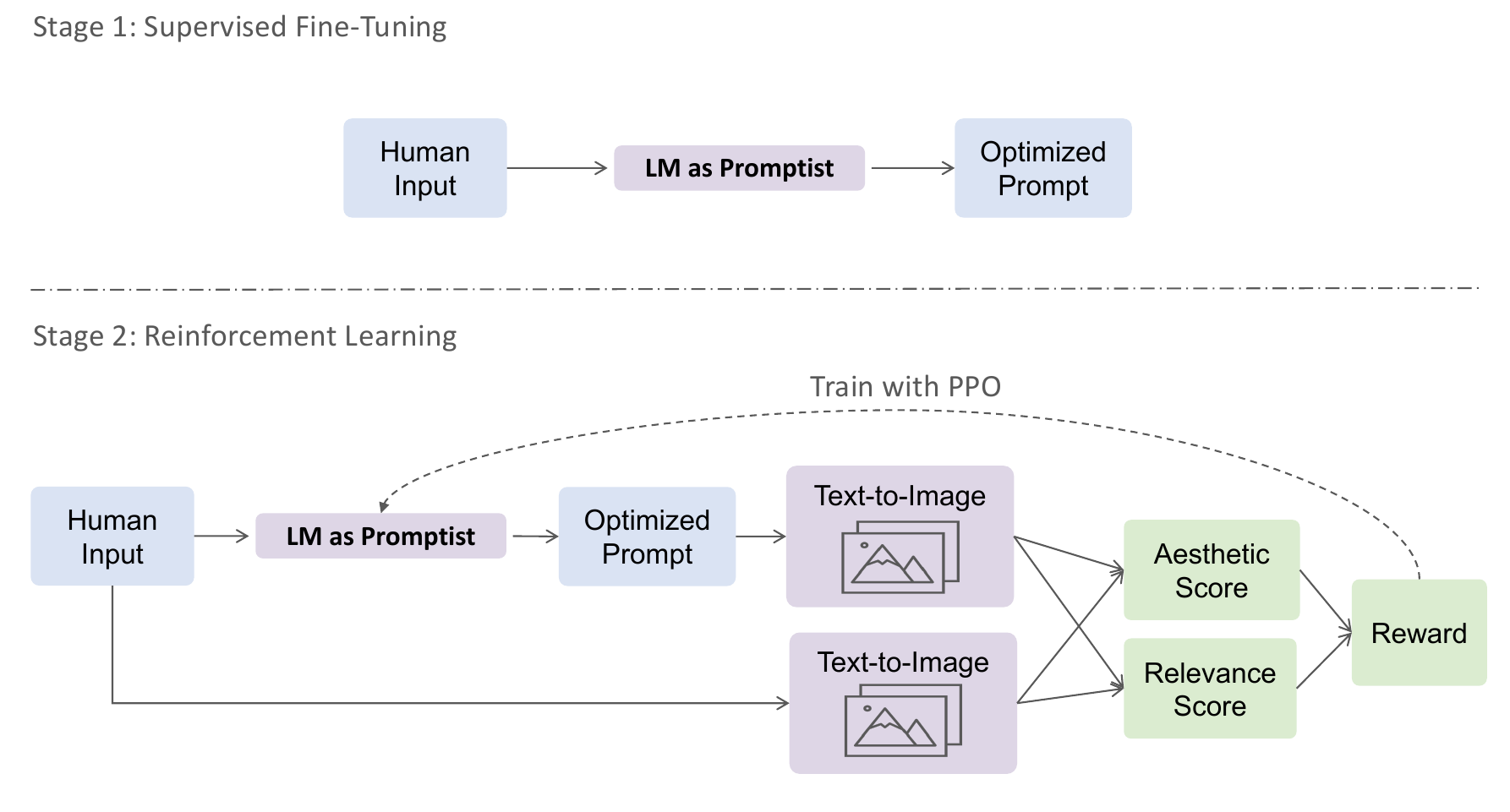}
\caption{
Overview of \our{} training: (1) supervised fine-tuning (SFT) on manually engineered prompts; (2) reinforcement learning (RL) to increase the rewards of generated images after prompt optimization.
}
\label{fig:method}
\end{figure*}

In this work, we propose a prompt adaptation framework for automatic prompt engineering via reinforcement learning.
Specifically, we first perform supervised fine-tuning with a pretrained language model (e.g., GPT) on a small collection of manually engineered prompts.
The finetuned model is used to initialize the prompt policy network for reinforcement learning.
Next, the model is trained by exploring optimized prompts of user inputs, where diverse beam search~\citep{dbs} is used to ensure generation quality and diversity.
The training objective is to maximize the reward, which is defined as a combination of relevance scores and aesthetic scores of generated images.
The relevance score reflects how much the original user intentions are retained after prompt adaptation.
The aesthetic score indicates what degree the generated images are aesthetically pleasing.

We conduct experiments with the publicly available Stable Diffusion models~\citep{sdiffusion}.
We evaluate our method using both the automatic reward metric and human preference ratings.
Experimental results show that the optimized prompts outperform human-engineered ones and the original inputs.
Human preference ratings also show consistent improvements across in-domain and out-of-domain prompts.
Moreover, we find that reinforcement learning is more favorable than supervised fine-tuning, especially on out-of-domain user inputs.
Overall, we show that language models can serve as a prompt interface that optimizes user input into model-preferred prompts.

Our contributions are as follows:
\begin{itemize}
\item We propose a general prompt optimization framework that adapts user input to model-preferred prompts.
\item We collect user queries and conduct extensive experiments on text-to-image generation.
\item Experimental results show that our method outperforms manual prompt engineering in terms of both automatic metrics and human preference ratings.
\end{itemize}

\section{Methods}
\label{sec:methods}

The goal of our prompt adaptation framework is to automatically perform prompt engineering.
Given user input of the text-to-image generator, our model learns to generate model-preferred prompts that obtain better output images while preserving their original intentions.
Figure~\ref{fig:method} presents the overview of our method.
The prompt optimization model is named \our{}, which is built upon a pretrained language model, such as GPT~\citep{gpt3}.
We first collect a set of human-engineered examples and use them to conduct supervised fine-tuning (Section~\ref{sec:sft}).
Next, we perform reinforcement learning (Section~\ref{sec:rl}) to maximize the target reward (Section~\ref{sec:reward}), which improves both relevance and quality of generated images.

\subsection{Supervised fine-tuning}
\label{sec:sft}

Initialized with a pretrained generative language model, the policy model is first finetuned on a set of prompt pairs before reinforcement learning.
A parallel prompt corpus $\train = \{ (\vx, \vy) \}$ contains prompt pairs of original user inputs $\vx$ and manually engineered examples $\vy$.
The training objective is to maximize the log-likelihood with teacher forcing:
\begin{align}
\Ls_\text{SFT} = - \mathbb{E}_{(\vx, \vy) \sim \train} \log p(\vy | \vx)
\end{align}
where the finetuned weights are used to initialize the policy network in reinforcement learning.

\paragraph{Collect human demonstrations}
We collect human-engineered prompts from Lexica\footnote{\url{https://lexica.art}}.
Most prompts are composed of two parts, i.e., main content that describes the user's intention, and some modifiers that customize the art style, such as artist names, and popular elements.
We use the crawled human-engineered prompts as targets.
In order to obtain parallel data, we use three methods to construct their source inputs.
First, we extract the main contents by trimming the modifiers and regard them as original user inputs.
Second, we randomly remove or shuffle some modifiers and use the remaining texts as source inputs.
Third, we use the OpenAI API \texttt{text-davinci-002} to rephrase the main contents and the human-engineered prompts, respectively. We find that the template ``\texttt{[Input] Rephrase:}'' works well in practice and translates input to a more user-friendly version.
As shown in Table~\ref{tbl:source_prompt}, given a target prompt ``\texttt{dieselpunk blue wolf with fuzzy tail, concept art, dramatic, fantasy, pixiv}'', there are four source prompts collected.

\begin{table}[h]
\caption{An example of a human-engineered prompt and four types of our constructed source prompts.}
\centering
\scalebox{0.87}{
\begin{tabular}{l l}
\toprule
\textbf{Human-engineered target prompt} & dieselpunk blue wolf with fuzzy tail, concept art, dramatic, \\ 
& fantasy, pixiv \\ \midrule
\textbf{Main content} & dieselpunk blue wolf with fuzzy tail \\ \midrule
\textbf{Main content with random modifiers} & dieselpunk blue wolf with fuzzy tail, dramatic \\ \midrule
\textbf{Rephrasing of main content} & A blue wolf with a fuzzy tail that looks like it belongs in a \\
& dieselpunk setting. \\ \midrule
\textbf{Rephrasing of target prompt} & {This is a dieselpunk-style blue wolf with a fuzzy tail. It looks} \\
 & {like it could be from a fantasy or dramatic piece of artwork.} \\
\bottomrule
\end{tabular}
}
\label{tbl:source_prompt}
\end{table}

\subsection{Reward definition}
\label{sec:reward}

We measure the quality of optimized prompts from two aspects, namely relevance and aesthetics.
The goal motivates us to define the reward function $\mathcal{R}(\cdot)$ from the above two perspectives.

First, we measure whether the generated images are relevant to the original input prompt after prompt adaptation.
To be specific, we first sample images by the text-to-image model conditioned on the optimized prompt, respectively. Then, we compute CLIP~\citep{clip} similarity scores to measure how relevant the generated images and the original input prompts are. The resulting relevance score is defined as:
\begin{align}
f_\text{rel}(\vx, &\vy) = \mathbb{E}_{i_\vy \sim \mathcal{G}(\vy)} [ f_\text{rel}(\vx, i_\vy) ] \\
f_\text{rel}(\vx, i_\vy) = &min(20*g_\text{CLIP}(\vx, i_\vy)-5.6, 0)
\end{align}
where $i_\vy \sim \mathcal{G}(\vy)$ means sampling images $i_\vy$ from the text-to-image model $\mathcal{G}$ with $\vy$ as input prompt, and $g_\text{CLIP}(\cdot,\cdot)$ stands for the CLIP similarity function. Notice that we always compute the similarity between the generated images and the original input prompt, which ensures the relevance score reflects the user preferences. 
We determine the specific form of the relevance score according to the approximate range of the clip score.
Experiments show that this form works well in reinforcement learning.
If the relevance score is relatively reasonable (larger than 0.28), we encourage the model to generate more aesthetically pleasing images.

Second, we employ the aesthetic predictor\footnote{\url{https://github.com/christophschuhmann/improved-aesthetic-predictor}} to quantify aesthetic preferences.
The predictor builds a linear estimator on top of a frozen CLIP model, which is trained by human ratings in the Aesthetic Visual Analysis~\citep{murray2012ava} dataset.
The aesthetic score is defined as:
\begin{align}
f_\text{aes}(\vx, \vy) = \mathbb{E}_{i_\vx \sim \mathcal{G}(\vx), i_\vy \sim \mathcal{G}(\vy)} [g_\text{aes}(i_\vy)\!-\!g_\text{aes}(i_\vx)]
\end{align}
where $g_\text{aes}(\cdot)$ denotes the aesthetic predictor, and $i_\vy, i_\vx$ are the images generated by the prompts $\vy$ and $\vx$, respectively. Notice that both $g_\text{CLIP}(\cdot)$ and $g_\text{aes}(\cdot)$ require the CLIP model, so we can share the CLIP forward pass during reward computation.

Finally, we define the overall reward by combining the above scores with an additional KL penalty, which is between the policy model $\pi_\vtheta$ and the supervised finetuned model $\pi_\text{SFT}$ with coefficient $\eta$:
\begin{align}
\label{eq:reward}
\begin{split}
\mathcal{R}(\vx, \vy) =&~ f_\text{aes}(\vx, \vy) + f_\text{rel}(\vx, \vy) \\
&-\eta~\text{log}{\frac{\pi_\vtheta(\vy|\vx)}{\pi_\text{SFT}(\vy|\vx)}}
\end{split}
\end{align}
The KL term is added to mitigate the overoptimization issue~\citep{instructgpt}.

\subsection{Reinforcement learning}
\label{sec:rl}

Starting from the supervised fine-tuning, we further finetune our model with reinforcement learning. We employ proximal policy optimization (PPO)~\citep{ppo}, which is empirically data-efficient and of reliable performance.
As a text generation problem, prompt optimization can be viewed as a Markov decision process (MDP) $\left\langle \mathcal{S}, \mathcal{A}, r, f_\text{st}, \gamma \right\rangle$ with a finite state space $\mathcal{S}$, action space $\mathcal{A}$, reward function $r$, state-transition probability function $f_\text{st}$, and a discount term $\gamma$. In an episode of prompt adaptation, the initial state $\vx \in \mathcal{S}$ is the input prompt with $n$ tokens $\vx = (x_1, \dots, x_n)$ where each token $x$ is from a finite vocabulary $\mathcal{V}$. At $t$-th time step, the agent selects an action $y_t \in \mathcal{V}$ according to the current policy model $y_t\sim\pi(y | \vx, \vy_{<t})$. With a deterministic state transition, the next state is $(\vx, \vy_{<t+1}) = (x_1, \dots, x_n, y_1, \dots, y_{t})$. The episode ends when the agent selects an end-of-sentence action.
The goal of the agent is to maximize the accumulated expected reward $\mathbb{E}_{\vx, \vy} \sum_t \gamma^t r(\vx, \vy_{<t}) = \mathbb{E}_{\vx, \vy} \mathcal{R}(\vx, \vy)$.

Let $\pi_\vtheta$ denote the policy model to be trained, we maximize the accumulated expected reward over a training set $\train' = \{ \vx \}$:
\begin{align}
    \mathcal{J} = \mathbb{E}_{\vx \sim \train', \vy \sim \pi_\vtheta} [\mathcal{R}(\vx, \vy)]
\end{align}
We implement both the policy model $\pi_\vtheta$ and the value function model as generative language models, with the language modeling head and the regression head, respectively.
The parameters of the two models are initialized from the supervised finetuned policy model $\pi_\text{SFT}$ and are optimized during reinforcement learning.
The supervised finetuned model $\pi_\text{SFT}$ and the score function model are frozen during training.
Besides, we employ the clipped probability ratios~\citep{ppo} to avoid large policy updates.

\section{Experiments}
\label{sec:exp}

We conduct experiments on public text-to-image model Stable Diffusion v1.4\footnote{\url{https://huggingface.co/CompVis/stable-diffusion-v1-4}} and v1.5\footnote{\url{https://huggingface.co/runwayml/stable-diffusion-v1-5}} . We use the DPM solver~\citep{dpmsolver} to accelerate image sampling and set the denoising steps to 20.

\subsection{Data collection}
For supervised fine-tuning, we collect 90k target prompts from Lexica website and construct four types of source prompts as described in Section~\ref{sec:sft}, obtaining 360k paired data in total.
At the reinforcement learning stage, we only require source prompts and the policy can explore better rephrasings itself.
We use three types of data: (1) in-domain prompts from DiffusionDB~\citep{diffusiondb}, which is a gallery of prompts specified by real users. We use the user input (main content) for exploration and the manually engineered prompt (with modifiers) for comparison, (2) out-of-domain image captions from COCO dataset~\citep{cocodata}, (3) image labels from ImageNet-21k~\citep{imagenet}, the sizes of which are 600k, 600k and 40k respectively.
We empirically observe that human-engineered prompts from Lexica perform better than those from DiffusionDB so we use the former in supervised fine-tuning.
To improve the data diversity, we add image caption data and image label data in reinforcement learning.
To avoid model bias in certain data formats, we randomize the capitalization of the first letter of each prompt and randomly add periods at the end of it.

\subsection{Settings}

For the policy model, we use GPT-2~\citep{gpt2} with 117M parameters, which is a multi-layer Transformer~\citep{transformer} decoder pretrained with causal language modeling.

\paragraph{Supervised fine-tuning}
We finetune GPT-2 to predict the target prompt conditioned on the source prompt with teacher forcing. 
The input format is \texttt{[Source] Rephrase:[Target]}. 
We use a batch size of 256, a learning rate of 5e-5, and a max length of 512. 
We finetune the model for 15k steps and choose a slightly underfitting checkpoint according to the validation loss which aims to avoid overfitting and provide a proper exploration space for the policy.

\begin{figure*}[t]
    \centering
    \includegraphics[width=0.99\textwidth]{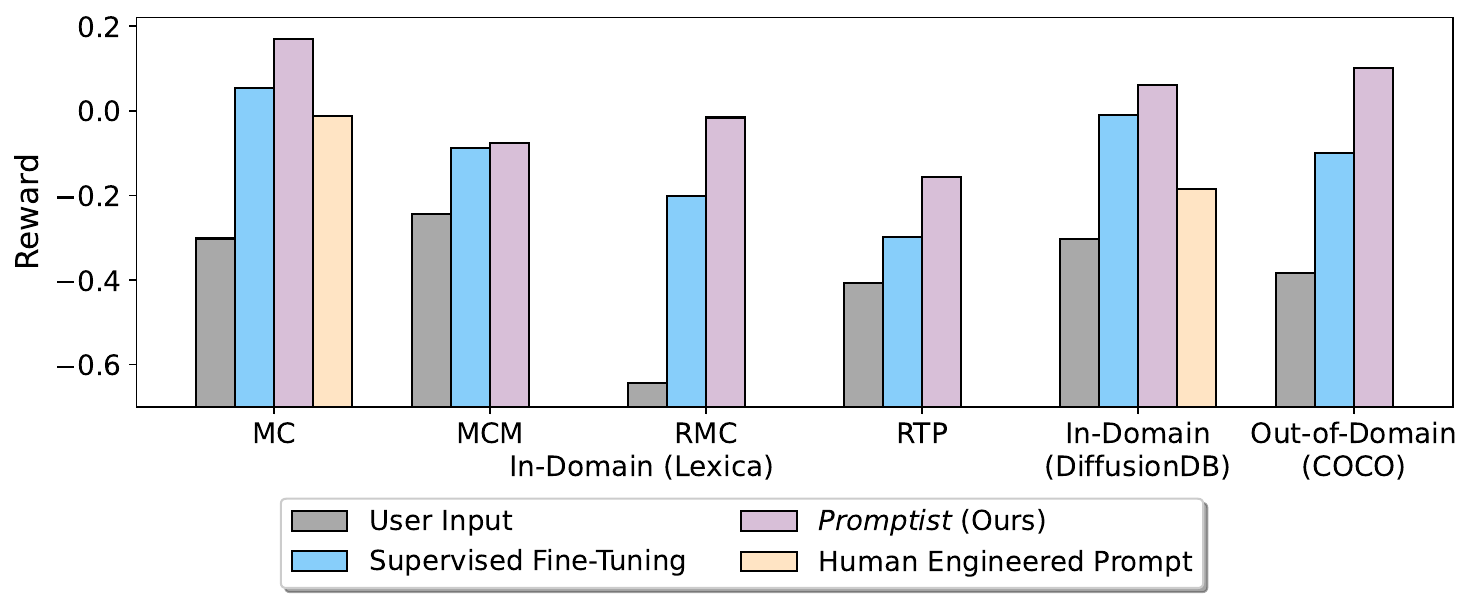}
    \caption{Reward comparison of optimized prompts with other baselines on in-domain and out-of-domain data. For in-domain Lexica prompts, we evaluate on four augmentations: main content (MC), main content with random modifiers (MCM), rephrasing of main content (RMC) and rephrasing of target prompt (RTP). Results indicate that the text-to-image model benefits a lot from our method.}
    \label{fig:aes_res}
\end{figure*}

\begin{table*}[tp!]
\caption{Absolute reward improvements of supervised fine-tuning and reinforcement learning. It is observed that RL generally outperforms the SFT-only model.}
\centering
\begin{tabular}{l c c c c c c}
\toprule
 & \multicolumn{4}{c}{In-Domain (Lexica)} & In-Domain & Out-of-Domain \\
 & MC & MCM & RMC & RTP & (DiffusionDB) & (COCO) \\
 \cmidrule(r){2-5} \cmidrule(l){6-6} \cmidrule(l){7-7} 
 SFT & 0.36 & 0.16 & 0.44 & 0.11 & 0.29 & 0.28 \\
 RL & 0.47 & 0.17 & 0.63 & 0.25 & 0.36 & 0.48 \\
 Gain & \pmb{+31\%} & \pmb{+6\%} & \pmb{+43\%} & \pmb{+127\%} & \pmb{+24\%} & \pmb{+71\%} \\
\bottomrule
\end{tabular}
\label{tbl:aesthetic_ratio}
\end{table*}

\paragraph{Reinforcement learning}
We train the policy with Proximal Policy Optimization~\citep[PPO]{ppo}.
The value and policy network are initialized from the supervised finetuned model.
The parameters of the value function are separated from the policy to avoid excessive competition between two objectives.
To guarantee the quality and diversity of exploration, we adopt diverse beam search~\citep{dbs} with a beam size of 8 and a diversity penalty of 1.0.
We find that having too long rephrasings occasionally produces aesthetically pleasing but misleading results, especially for short user input like image labels.
In order to prevent the model from only exploring long completions, the maximum generation length at each step is set to a random value from 15 to 75 so that the policy can learn to adjust the generation length for each prompt.
We randomly choose one of the returned completions after diverse beam search to update the policy.
We generate three images per prompt and compute the average reward to reduce variance.
We train the policy for 12k episodes, four PPO epochs per batch with one minibatch each, with a batch size of 256 and a constant learning rate of 5e-5.
The value loss coefficient and the KL reward coefficient are kept at 2.3 and 0.2 respectively.
We do not cherry-pick checkpoints and directly use the final checkpoint for evaluation.
Please refer to the Appendix~\ref{app:hyperparameter} for more training details and Appendix~\ref{app:budget} for computational resources.

\paragraph{Evaluation}
In order to evaluate how text-to-image models benefit from the prompt adaptation, we compare the reward value computed by two automatic predictors (Section~\ref{sec:reward}).
Moreover, we use human preference ratings to demonstrate real user feedback.
We adopt beam search with a beam size of 8 and a length penalty of 1.0.
We evaluate our method on held-out data from training distribution, including in-domain data from Lexica with four augmentations, in-domain data from DiffusionDB, and out-of-domain COCO data.
Each category contains 256 prompts.
In-domain data has corresponding manually engineered prompts for comparison, and the out-of-domain data is used to verify whether our method can generalize to new domains.
Except for the user input and manually engineered baseline, we also consider the supervised finetuned model as a baseline that can reflect the importance of reinforcement learning.

\subsection{Results}
\begin{table*}[ht!]
\caption{Images generated by user input and optimized prompts using Stable Diffusion v1.4. Each group contains three images generated with three different random seeds. 
We observe that optimized prompts can generate more aesthetically pleasing images than original user input.}
\centering
\begin{tabular}{p{0.45\textwidth} p{0.45\textwidth}}
\toprule
\textbf{User Input}
& \textbf{Optimized Prompt} \\
\midrule
{A rabbit is wearing a space suit}
& \multicolumn{1}{m{0.49\textwidth}}{A rabbit is wearing a space suit, digital Art, Greg rutkowski, Trending cinematographic artstation} \\
\midrule
{\begin{minipage}[t]{.48 \textwidth}
\includegraphics[width=.3\textwidth]{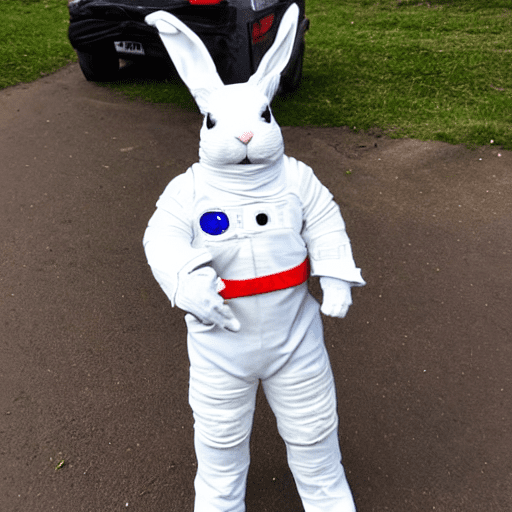}
\includegraphics[width=.3\textwidth]{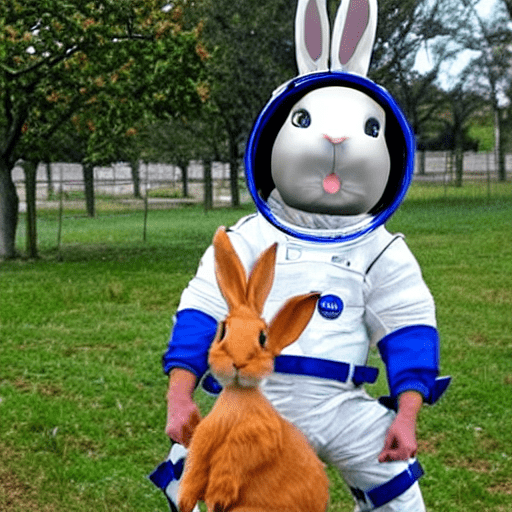}
\includegraphics[width=.3\textwidth]{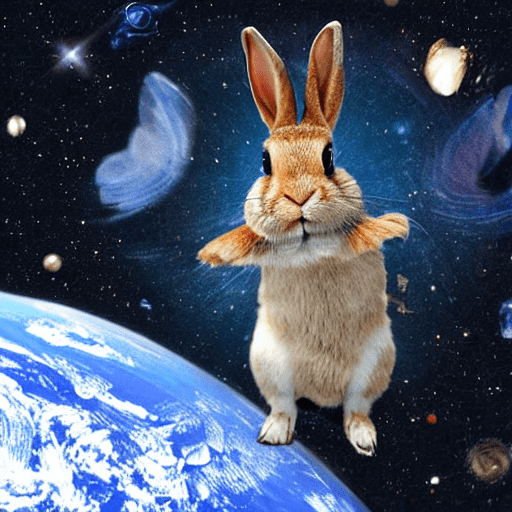}
\end{minipage}}
& \quad {\begin{minipage}[t]{.48 \textwidth}
\includegraphics[width=.3\textwidth]{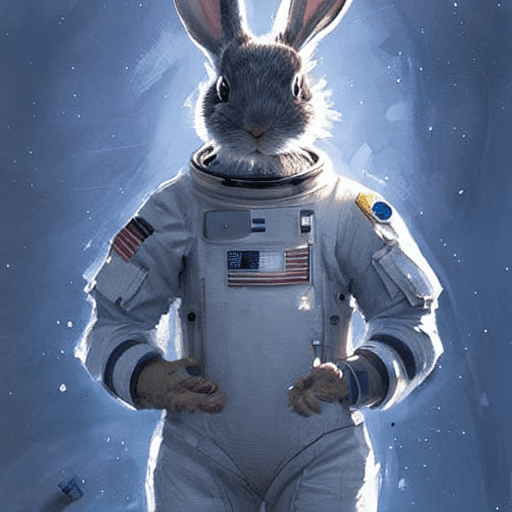}
\includegraphics[width=.3\textwidth]{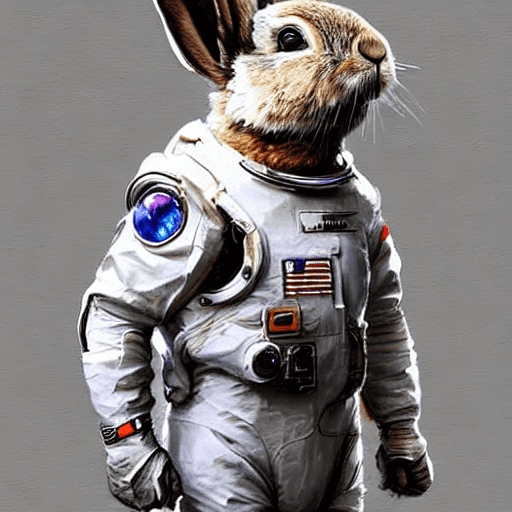}
\includegraphics[width=.3\textwidth]{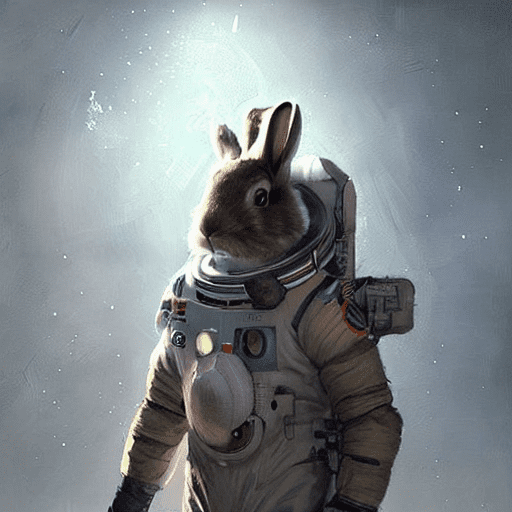}
\end{minipage}} \\
\midrule
{Several railroad tracks with one train passing by}
& \multicolumn{1}{m{0.49\textwidth}}{several railroad tracks with one train passing by, hyperdetailed, artstation, cgsociety, 8 k} \\
\midrule
{\begin{minipage}[t]{.48 \textwidth}
\includegraphics[width=.3\textwidth]{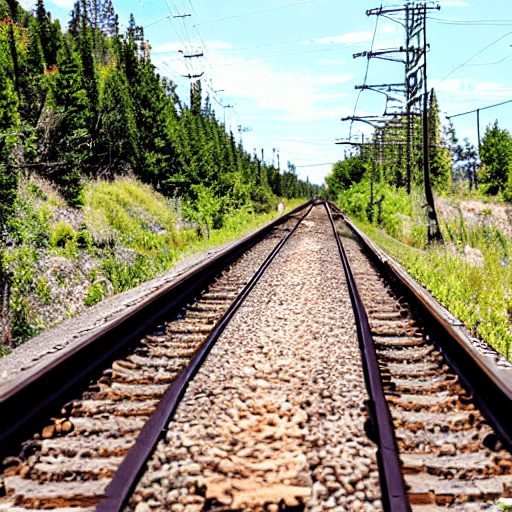}
\includegraphics[width=.3\textwidth]{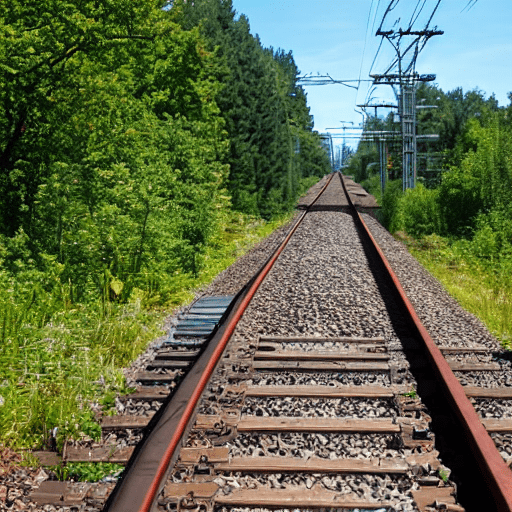}
\includegraphics[width=.3\textwidth]{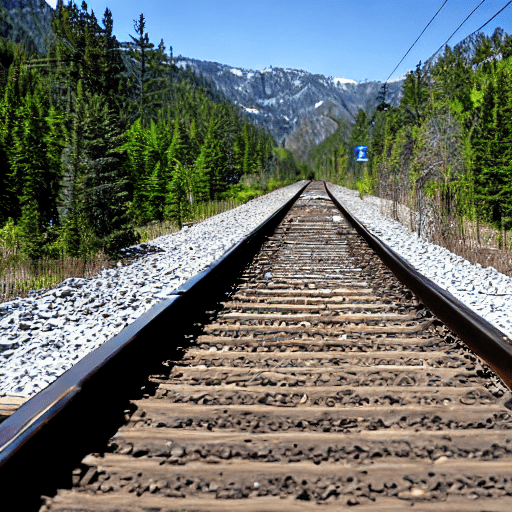}
\end{minipage}}
& \quad {\begin{minipage}[t]{.48 \textwidth}
\includegraphics[width=.3\textwidth]{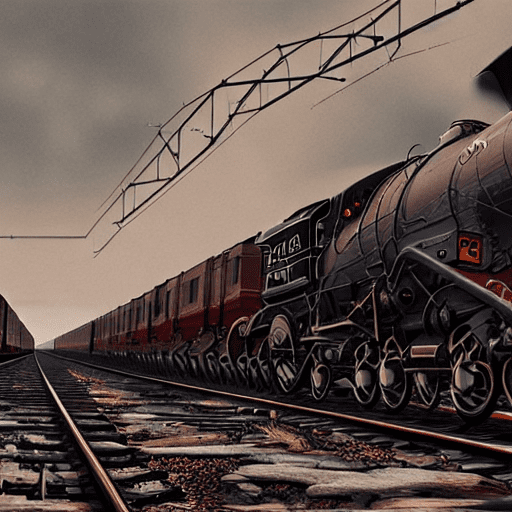}
\includegraphics[width=.3\textwidth]{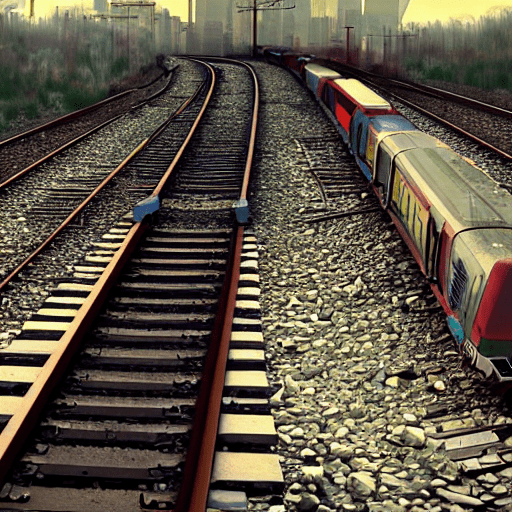}
\includegraphics[width=.3\textwidth]{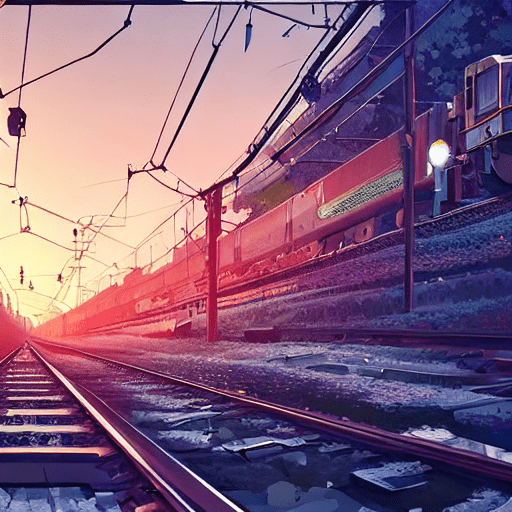}
\end{minipage}} \\
\midrule
{The roof is wet from the rain}
& \multicolumn{1}{m{0.49\textwidth}}{the roof is wet from the rain, intricate, elegant, highly detailed, digital painting, artstation, concept art, smooth, sharp focus, illustration,} \\
\midrule
{\begin{minipage}[t]{.48 \textwidth}
\includegraphics[width=.3\textwidth]{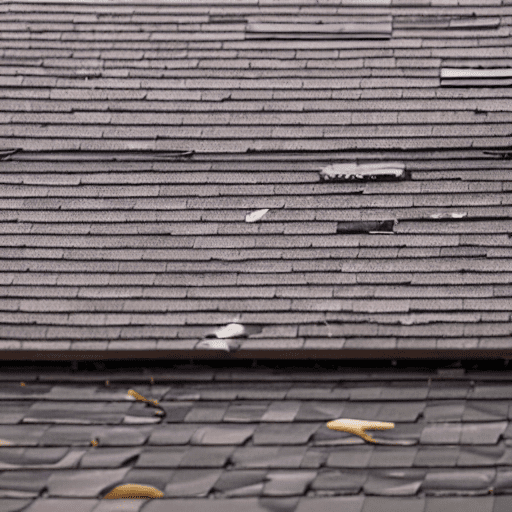}
\includegraphics[width=.3\textwidth]{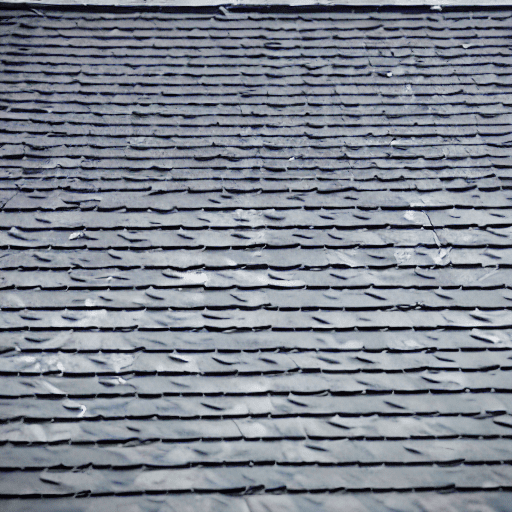}
\includegraphics[width=.3\textwidth]{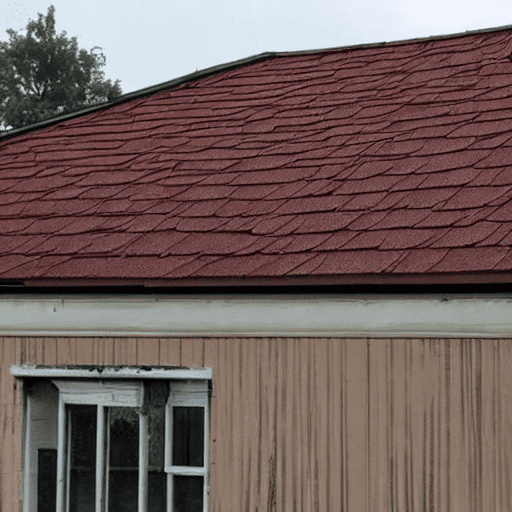}
\end{minipage}}
& \quad {\begin{minipage}[t]{.48 \textwidth}
\includegraphics[width=.3\textwidth]{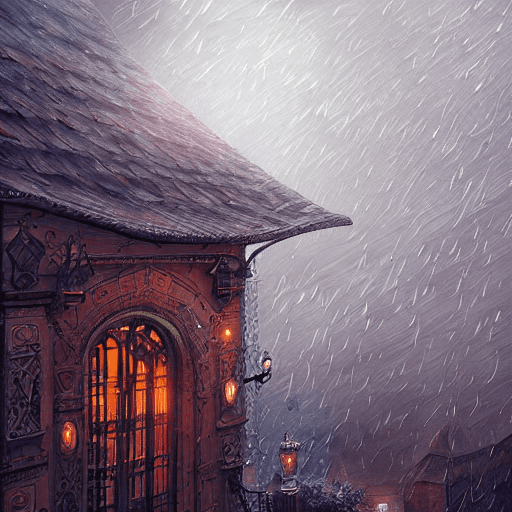}
\includegraphics[width=.3\textwidth]{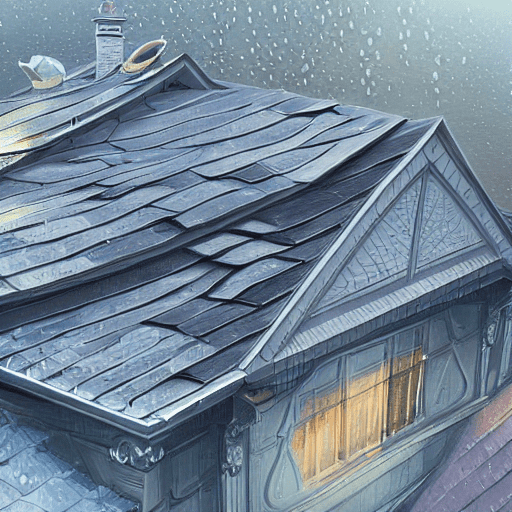}
\includegraphics[width=.3\textwidth]{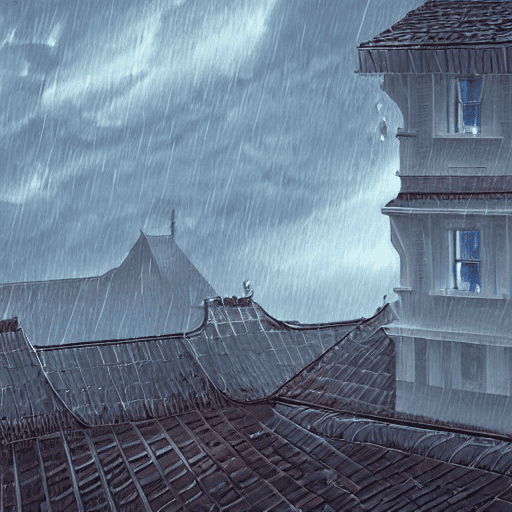}
\end{minipage}} \\
\midrule
{Cats dancing in a space club}
& \multicolumn{1}{m{0.49\textwidth}}{Cats dancing in a space club, digital painting, artstation, concept art, soft light, hdri, smooth, sharp focus, illustration, fantasy,} \\
\midrule
{\begin{minipage}[t]{.48 \textwidth}
\includegraphics[width=.3\textwidth]{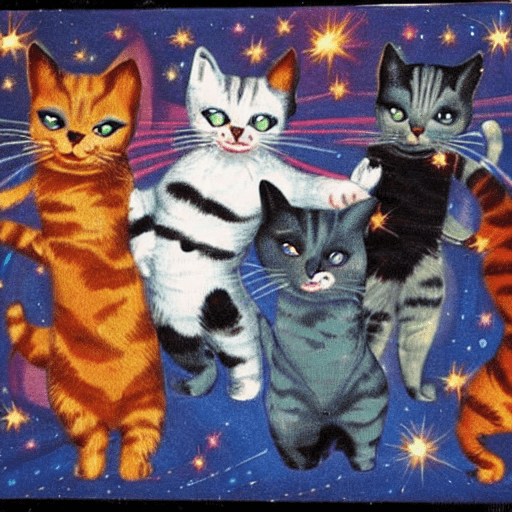}
\includegraphics[width=.3\textwidth]{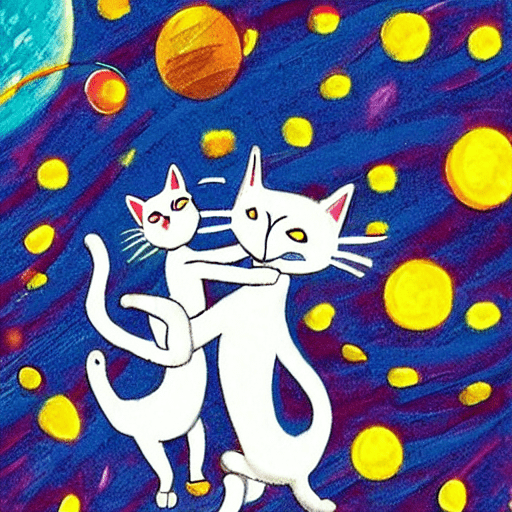}
\includegraphics[width=.3\textwidth]{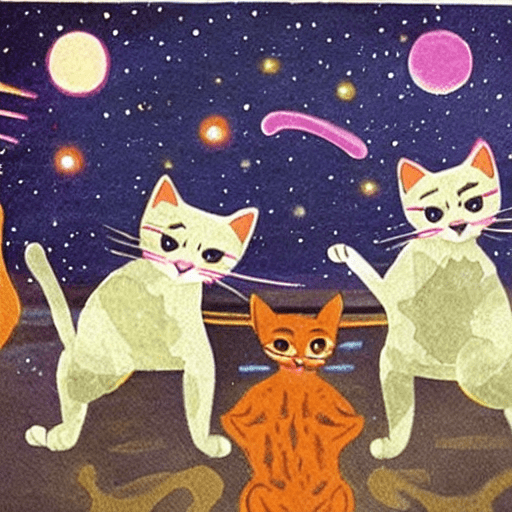}
\end{minipage}}
& \quad {\begin{minipage}[t]{.48 \textwidth}
\includegraphics[width=.3\textwidth]{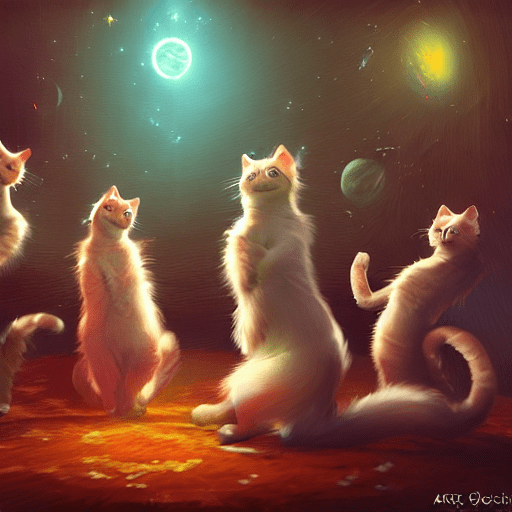}
\includegraphics[width=.3\textwidth]{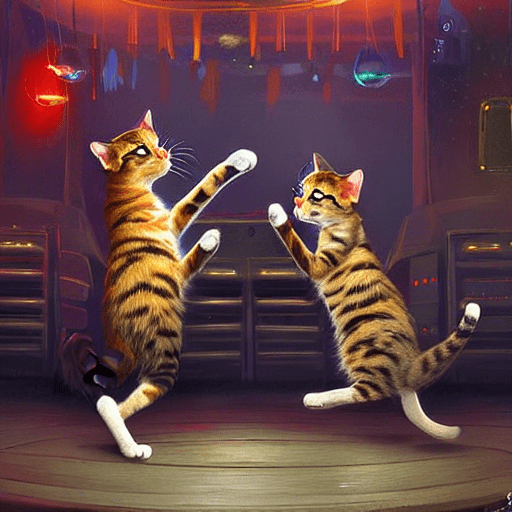}
\includegraphics[width=.3\textwidth]{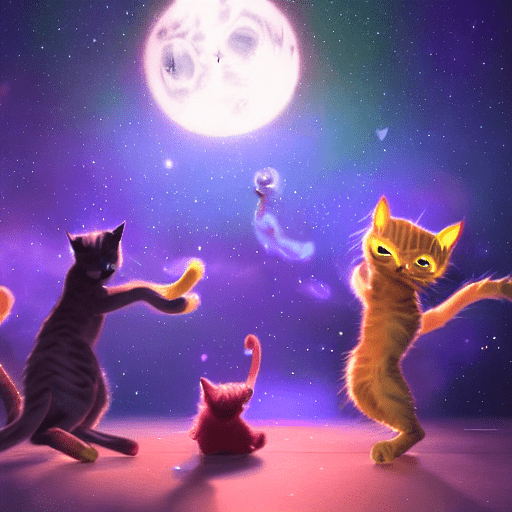}
\end{minipage}} \\
\bottomrule
\end{tabular}
\label{tbl:text_image_pair}
\end{table*}

\paragraph{Optimized prompts obtain higher reward improvements than manual engineering.}
We evaluate optimized prompts on held-out data by generating three images for each prompt and computing the average reward value. Figure~\ref{fig:aes_res} shows that the reward value can be improved regardless of the engineering method, which suggests the misalignment problem between user-friendly prompts and model-preferred prompts is serious.
Compared with the strong baseline of manually engineered prompts, optimized prompts can still achieve considerable reward improvements.
Furthermore, optimized prompts perform even better on rephrased versions (i.e., RMC, and RTP), and out-of-domain data.
These prompts are more user-friendly but cause more significant reward drops on generation results, especially on the rephrasing of the main content.
Benefiting from automatic prompt engineering, optimized prompts can align well between two different domains from users and text-to-image models respectively.

\begin{table}[h]
\caption{Evaluation of the aesthetic score and relevance score on DiffusionDB.}
\centering
\begin{tabular}{lcc}
\toprule
 & \textbf{Aesthetic} & \textbf{Relevance} \\ \midrule
User Input & 5.47 & 0.28 \\ \midrule
Human Engineered Prompt & 5.87 & \pmb{0.26} \\
Supervised Fine-tuning & 6.15 & 0.25 \\
\our (Ours) & \pmb{6.26} & \pmb{0.26} \\
\bottomrule
\end{tabular}
\label{tbl:aes_rel_score}
\end{table}
We also present the evaluation results of the aesthetic score and relevance score respectively in Table~\ref{tbl:aes_rel_score}.
We empirically found that the generated images are relevant enough to the input prompt if the relevance score (CLIP score) is around 0.26. 
As mentioned at Section~\ref{sec:reward}, we design the reward function which encourages the model to generate more aesthetically pleasing images if the relevance score is good enough. On the DiffusionDB dataset, our RL method improves the SFT baseline in terms of relevance score from 0.25 to 0.26, and the human-engineered baseline also obtains a relevance score of 0.26.
Moreover, the aesthetic score of our model is improved significantly over both the human-engineered prompts and the supervised fine-tuned model.
It demonstrates that our method generates images with good relevance and much better aesthetic scores.

We provide some images generated by user input and its corresponding optimized prompt in Table~\ref{tbl:text_image_pair}.
Each group consists of three images generated by different random seeds.
We observe that images generated by user input are intuitively uninspiring while optimized prompts can not only retain the original intentions but also induce the model to produce more remarkable results.
For example, generated images are crude when prompted with ``A rabbit is wearing a space suit''.
After prompt optimization, generated images become more bright and more expressive.

\begin{table*}[tp!]
\caption{Human evaluation results. The different colors represent how many images generated by corresponding prompts are considered more aesthetically pleasing. The {\color{Orange} orange} block means that both prompts produce equally pleasing images.}
\centering
\bgroup
\def\arraystretch{1.9}
\begin{tabular}{l c c}
\toprule
 & {\color{RoyalBlue} Optimized} vs. {\color{Gray} User Input} & {\color{RoyalBlue} Optimized} vs. {\color{Gray} Manually Engineered} \\
 \midrule
 In-Domain & \begin{minipage}{.36\textwidth}\includegraphics[width=.98\textwidth]{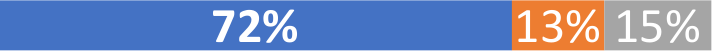}\end{minipage} & \begin{minipage}{.36\textwidth}\includegraphics[width=.98\textwidth]{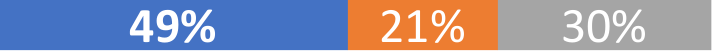}\end{minipage} \\
 \hfill
 Out-of-Domain & \begin{minipage}{.36\textwidth}\includegraphics[width=.98\textwidth]{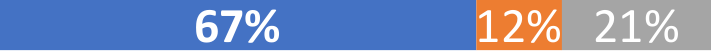}\end{minipage} & ------ \\
\bottomrule
\end{tabular}
\egroup
\label{tbl:human_eval}
\end{table*}

\begin{figure*}[t]
\centering
\includegraphics[width=0.99\textwidth]{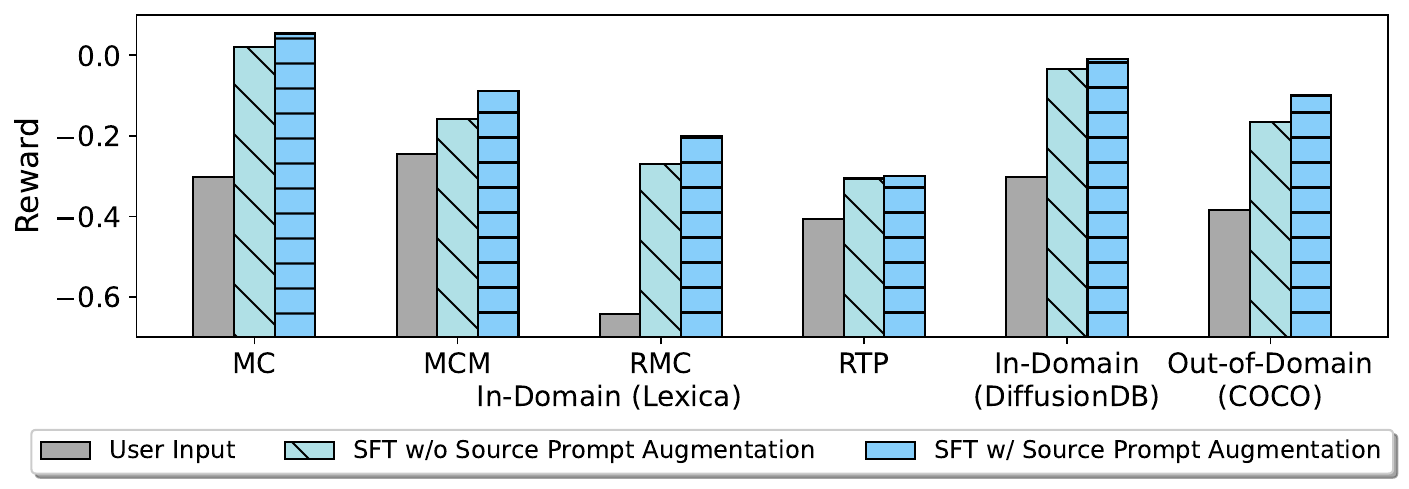}
\caption{
Reward comparison of supervised fine-tuning with or without source prompt augmentation. For in-domain Lexica prompts, we evaluate on four augmentations: main content (MC), main content with random modifiers (MCM), rephrasing of main content (RMC) and rephrasing of target prompt (RTP). It is observed that source prompt augmentation in supervised fine-tuning can boost performance on both in-domain data and out-of-domain data.
}
\label{fig:ablation}
\end{figure*}

\paragraph{Reinforcement learning can further boost the reward value.}
Reinforcement learning in our method is supposed to perform better on out-of-domain data through explorations.
To quantify its effect, we compute the ratio of reward improvements after fine-tuning and reinforcement learning.
As shown in Table~\ref{tbl:aesthetic_ratio}, reinforcement learning brings 31\%, 24\%, and 71\% average improvements on in-domain main content from Lexica, DiffusionDB, and out-of-domain COCO data.
In-domain prompts are very similar to the data we used in supervised fine-tuning, so reward improvements are relatively saturated in the first stage and improvements of reinforcement learning on them are correspondingly smaller. 
Oppositely, out-of-domain data such as COCO captions are more similar to user input and unseen during the first stage.
The policy must learn to adapt better to new domains through exploration, so their improvements on these prompts are more prominent.
Surprisingly, although in-domain Lexica prompts and their augmentations are not used, reinforcement learning still exhibits better generalization capability on them.
The boost is remarkable on those prompts that fine-tuning cannot optimize well (43\% on rephrasings of main content and 127\% on rephrasings of target prompt).
These results suggest that given appropriate human queries, reinforcement learning can optimize them to adapt to different domains and boost reward improvements.

To further demonstrate the effectiveness of our framework, we also present the results of our model on Stable Diffusion v1.5 in Appendix~\ref{app:sdv1.5}, comparisons with the heuristic baseline in Appendix~\ref{app:heuristic_baseline} and the results on different categories and lengths of prompts in Appendix~\ref{app:category_length}.

\subsection{Human evaluation}

The reward function of our model is defined by two automatic metrics, aesthetic score and relevance score predicted by neural networks, which may have some discrepancies from real human feedback.
Therefore, we additionally evaluate whether optimized prompts actually make humans more satisfied.
We generate two images for each user input and the optimized prompt.
Afterward, three held-out annotators are asked to rank the two groups of images in preference order and we compute the average preference distribution.
Results are shown in Table~\ref{tbl:human_eval}.
We observe that annotators generally prefer images generated by optimized prompts over their original input.
Compared with manually engineered prompts, optimized prompts yield less gain over user input.
It suggests that the aesthetic score can measure the quality of generated images to some extent, it would be better if direct human feedback is included in the reward function.

\subsection{Ablation of source prompt augmentation}
As described in Section~\ref{sec:sft}, we crawl human-engineered prompts as target prompts and use the main content without any modifiers as source prompts.
To enable the supervised fine-tuned model to generalize better on unseen domains, we propose different augmentation methods that improve the diversity of source prompts.
We compare the fine-tuning performance with and without the augmentation strategy and results are shown in Figure~\ref{fig:ablation}.
We observe that fine-tuning with source prompt augmentation brings consistent improvements on both in-domain held-out data and out-of-domain data.
From the results on MCM, adding some random modifiers to the user input slightly obtains reward improvement but it is not as distinct as the improvement brought by fine-tuning, indicating that we should customize modifiers for each individual prompt and automatic prompt engineering is a promising way to tackle it.
Compared with other data, prompts rephrased by \texttt{text-davinci-002} are more difficult to optimize at the fine-tuning stage and they benefit more from reinforcement learning.
Overall, source prompt augmentation makes the fine-tuned model generalize better and is important in our prompt adaptation framework.

\section{Related work}

\paragraph{Prompt engineering.}

Manual prompt engineering is a natural way to optimize prompts. Manually designed cloze-style prompts have been used to probe knowledge from pre-trained language models~\citep{petroni2019language,dai2022knowledge}. In addition to knowledge probing, models are also prompted to handle NLP tasks with manually designed prefix prompts~\citep{gpt3,glam}.
Recent work has explored how to write prompts to improve performance~\citep{wei2022chain}. Despite the success of manually-crafted prompts, designing prompts takes time and experience~\citep{shin-etal-2021-constrained} and can be sub-optimal~\citep{jiang2020how}. In particular, when using text-to-image models, users have to carefully select and compose sentences to achieve a certain visual style~\citep{promptguideline,promptmodifier,guidebook}.
Thus, various methods focus on automatically searching prompts by mining~\citep{jiang2020how}, paraphrasing~\citep{haviv-etal-2021-bertese}, and text generation~\citep{gao-etal-2021-making}. Besides, continuous prompt methods treat the prompts as additional continuous parameters of pre-trained models and directly optimize the parameters on downstream tasks~\citep{li-liang-2021-prefix,tsimpoukelli2021multimodal,zhou2022learning}. However, continuous prompt methods require access to manipulating the model, and the learned prompts lack interpretability. In contrast, our methods directly optimize prompts in text format, which can fit in black-box downstream systems such as text-to-image models.

\paragraph{Learning from human feedback.} Our work is related to research on learning from human feedback, which has been widely studied in machine learning problems.
Several studies propose to continually improve dialogue systems by collecting human feedback after deployment~\citep{hancock-etal-2019-learning,shuster2020deploying,xu2022learning}. Besides, human feedback has also been also applied to human-in-the-loop methods for entity linking~\citep{klie-etal-2020-zero}, semantic parsing~\citep{yao-etal-2019-model}, etc.
Recent research on reinforcement learning from human feedback (RLHF) has shown promising results on machine learning problems, ranging from classical RL tasks~\citep{christiano2017deep,ibarz2018reward} to a wide range of natural language processing tasks, including text summarization~\citep{learningtosummarize,ziegler2019fine}, dialogue~\citep{jaques2019way}, and general text generation tasks~\citep{instructgpt}. Differently, our goal is to automatically optimize prompts for text-to-image models.

\paragraph{Text-to-image models.}
Text-to-image synthesis models are typically trained to generate images conditioned on text. Text-to-image synthesis has been widely studied using GANs~\citep{reed2016generative,reed2016learning,tao2022df}. More recently, text-to-image models are further improved with large-scale auto-regressive models~\citep{ramesh2021zero,ding2021cogview} or diffusion-based models~\citep{sdiffusion,gu2022vector}.

\section{Conclusion}

We propose to automatically optimize prompts for text-to-image models so that the user input and model-preferred prompts can be well aligned.
We evaluate our method with Stable Diffusion.
Experimental results show that prompt adaptation outperforms human prompt engineering and supervised fine-tuning, in terms of automatic metrics and human evaluation.
The exploration nature of reinforcement learning enables the model to go beyond teacher forcing, which improves generalization over out-of-domain examples.
The proposed method is flexible to align human intentions and model-favored languages.
Although our experiments are conducted on text-to-image models, the framework can be easily applied to other tasks for prompt adaptation.
Rather than using automatic score functions as rewards, we can directly use human feedback as supervision to train a reward model~\citep{instructgpt}.
Moreover, using a larger-size language model as the prompt interface tends to improve the optimization quality.

\section*{Limitations}
We crawl human-engineered prompts from the Lexica website as golden prompts to guide the supervised fine-tuning process.
The crawled prompts contain some biases.
For example, we observe that they tend to generate more artwork instead of realistic photographs because most of them contain one or more artist names.
Besides, the proportion of prompts about portraits is relatively higher than those about other categories. 
Although the reinforcement learning stage can mitigate these issues, it would be better to balance the art styles and objects at the beginning.
Moreover, we currently only apply our framework to text-to-image models.
As the proposed framework is general to prompt-guided generation,
we will apply it to other generative models like text-only models and text-to-video models for future work.

\begin{ack}
We would like to thank Tan Yan for the helpful discussions.
\end{ack}

\bibliography{promptist}
\bibliographystyle{plainnat}

\clearpage
\appendix
\section*{Appendix}
\section{Hyperparameter settings}
\label{app:hyperparameter}

\begin{table*}[h]
\caption{Hyperparameter settings of supervised fine-tuning (SFT) and reinforcement learning (RL).}
\centering
\begin{tabular}{lll}
\toprule
\textbf{Hyperparameters} & \textbf{SFT} & \textbf{RL} \\ \midrule
Batch Size & 256 & 256 \\
Learning Rate & 5e-5 & 5e-5 \\
Training Steps & 15000 & 12000 \\
Max Length & 512 & 512 \\
Dropout & 0.0 & 0.0 \\
Optimizer & Adam & Adam \\
Adam $\epsilon$ & 1e-6 & 1e-6 \\
Adam $\beta$ & (0.9, 0.999) & (0.9, 0.95) \\
Weight Decay & 0.1 & 1e-6 \\
\bottomrule
\end{tabular}
\end{table*}

\section{Computational budget}
\label{app:budget}
Our experiments are implemented on V100 (32GB) GPU.

\begin{table}[h]
\caption{Computational budget of supervised fine-tuning (SFT) and reinforcement learning (RL).}
\centering
\begin{tabular}{lll}
\toprule
 & \textbf{SFT} & \textbf{RL} \\ \midrule
The Number of GPUs & 4 & 32 \\
GPU Hours & 3 hours & 2.5 days \\
\bottomrule
\end{tabular}
\end{table}

\section{Results on Stable Diffusion v1.5.}
\label{app:sdv1.5}
\begin{table}[h]
\caption{Results on Stable Diffusion v1.5}
\centering
\begin{tabular}{lccc}
\toprule
 & \textbf{Lexica} & \textbf{DiffusionDB} & \textbf{COCO} \\ \midrule
User Input & -0.31 & -0.32 & -0.37 \\ \midrule
Human Engineered Prompt & -0.05 & -0.18 & - \\
Supervised Fine-tuning & -0.04 & -0.16 & -0.1 \\
\our (Ours) & \pmb{0.05} & \pmb{0.06} & \pmb{0.11} \\
\bottomrule
\end{tabular}
\end{table}

\newpage
\section{Comparisons with heuristic baseline}
\label{app:heuristic_baseline}

\begin{table}[h]
\caption{Combinations of common tags.}
\centering
\begin{tabular}{cc}
\toprule
 \textbf{Tag} & \textbf{Content} \\ \midrule
 1 & {artstation, highly detailed, elegant} \\ 
 2 & {8 k, trending on artstation, concept art} \\ 
 3 & {digital painting, intricate, fantasy} \\
 4 & {illustration, smooth, octane render} \\ 
 5 & {digital art, 8k, intricate} \\ 
 6 & {highly detailed, elegant, smooth} \\
\bottomrule
\end{tabular}
\label{tbl:commontag}
\end{table}

\begin{table}[h]
\caption{Comparisons with the heuristic baseline.}
\centering
\addtolength{\tabcolsep}{-1pt}
\begin{tabular}{lcccccccccc}
\toprule
\textbf{Data} & \textbf{User} & \textbf{Tag1} & \textbf{Tag2} & \textbf{Tag3} & \textbf{Tag4} & \textbf{Tag5} & \textbf{Tag6} & Human & SFT & Ours \\ \midrule
Lexica & -0.32 &	0.07 &	-0.06	 &0.06 &	-0.17 &	-0.05 &	-0.28 &	-0.02 &	0.03 &	\pmb{0.14} \\
DiffusionDB & -0.3 &	0 &	-0.07 &	-0.16 &	-0.1 &	-0.17 &	-0.31 &	-0.21 &	-0.01 &	\pmb{0.06} \\
COCO & -0.38 &	-0.24 &	-0.29 &	-0.2 &	-0.33 &	-0.32 &	-0.41 &	- &	-0.1 &	\pmb{0.1}
 \\
\bottomrule
\end{tabular}
\label{tbl:res_commontag}
\end{table}

To compare the performance of our proposed framework with the heuristic baseline, we select the top 15 frequent tags from human-engineered prompts and randomly combine them to create six groups of common tags. The specific tags are presented in Table~\ref{tbl:commontag}.
We concatenate the user input with these common tags and compute their reward. The results are in Table~\ref{tbl:res_commontag}.

While using these common tags can improve the reward to some extent, we found that their performance varies significantly across different domains. For instance, tag3 performs well on COCO and Lexica but poorly on DiffusionDB. It suggests that relying on a handful of common hand-selected tags may not be practical in real-world scenarios. In contrast, our proposed framework can perform well across domains and improve a lot over the common tags.

\newpage
\section{Results on different categories and lengths of prompts}
\label{app:category_length}
We aim to validate the effectiveness of our method on different categories and different lengths. 
In Figure~\ref{fig:aes_res}, we divide the prompts into several categories according to the prompt pattern, there are MC, MCM, RMC, RTP, in-domain DiffusionDB and out-of-domain COCO. Results show that optimized prompts are generally effective for all these categories. For semantic categories, these prompts have no clear boundaries. Therefore, we use RoBERTa-Large~\cite{roberta} to get the sentence embedding of each prompt and perform K-means clustering on these prompts and divide them into five categories. We list their proportion and their reward in Table.
For length ablation, we also classify them into five categories according to the length of input tokens. The results are in Table~\ref{tbl:length}.

We observe that the performance of different lengths and semantic categories varies slightly but our model can improve the reward generally. When conducting reinforcement learning, we build large-scale prompts from both in-domain data and out-of-domain data, which cover a wide range of prompts with different lengths and semantic categories.

\begin{table}[h]
\caption{Results on different semantic categories of prompts.}
\centering
\begin{tabular}{lccccc}
\toprule
 \textbf{Cluster} & \textbf{1} & \textbf{2} & \textbf{3} & \textbf{4} & \textbf{5} \\ \midrule
 Proportion & 0.18 &	0.08 &	0.17 &	0.21 &	0.36 \\ 
 User Input & -0.39 & -0.29 & -0.37 & -0.34 & -0.31 \\ 
 \our{} (Ours) & -0.01 &	0.04 &	0.13 &	0.06 &	0.1 \\
\bottomrule
\end{tabular}
\label{tbl:cluster}
\end{table}

\begin{table}[h]
\caption{Results on different lengths of prompts.}
\centering
\begin{tabular}{lccccc}
\toprule
 \textbf{Length} & \textbf{0$\sim$10} & \textbf{10$\sim$20} & \textbf{20$\sim$30} & \textbf{30$\sim$40} & \textbf{>40} \\ \midrule
 Proportion & 0.21 &	0.48 &	0.2 &	0.07 &	0.04 \\ 
 User Input & -0.48 &	-0.33 &	-0.18 &	-0.28 &	-0.22 \\ 
 \our{} (Ours) & -0.02 &	0.11 &	0.08 &	0.05 &	0.06 \\
\bottomrule
\end{tabular}
\label{tbl:length}
\end{table}

\end{document}

%% file: math_commands.tex
\usepackage{amsmath,amsfonts,bm}

\def\eqref#1{equation~\ref{#1}}

\def\1{\bm{1}}
\newcommand{\train}{\mathcal{D}}

\def\vtheta{{\bm{\theta}}}

\def\vx{{\bm{x}}}
\def\vy{{\bm{y}}}

\DeclareMathAlphabet{\mathsfit}{\encodingdefault}{\sfdefault}{m}{sl}
\SetMathAlphabet{\mathsfit}{bold}{\encodingdefault}{\sfdefault}{bx}{n}

\newcommand{\Ls}{\mathcal{L}}



%% file: main.bbl
\begin{thebibliography}{51}
\providecommand{\natexlab}[1]{#1}
\providecommand{\url}[1]{\texttt{#1}}
\expandafter\ifx\csname urlstyle\endcsname\relax
  \providecommand{\doi}[1]{doi: #1}\else
  \providecommand{\doi}{doi: \begingroup \urlstyle{rm}\Url}\fi

\bibitem[Brown et~al.(2020)Brown, Mann, Ryder, Subbiah, Kaplan, Dhariwal, Neelakantan, Shyam, Sastry, Askell, Agarwal, Herbert-Voss, Krueger, Henighan, Child, Ramesh, Ziegler, Wu, Winter, Hesse, Chen, Sigler, Litwin, Gray, Chess, Clark, Berner, McCandlish, Radford, Sutskever, and Amodei]{gpt3}
Tom Brown, Benjamin Mann, Nick Ryder, Melanie Subbiah, Jared~D Kaplan, Prafulla Dhariwal, Arvind Neelakantan, Pranav Shyam, Girish Sastry, Amanda Askell, Sandhini Agarwal, Ariel Herbert-Voss, Gretchen Krueger, Tom Henighan, Rewon Child, Aditya Ramesh, Daniel Ziegler, Jeffrey Wu, Clemens Winter, Chris Hesse, Mark Chen, Eric Sigler, Mateusz Litwin, Scott Gray, Benjamin Chess, Jack Clark, Christopher Berner, Sam McCandlish, Alec Radford, Ilya Sutskever, and Dario Amodei.
\newblock Language models are few-shot learners.
\newblock In \emph{Advances in Neural Information Processing Systems}, volume~33, pages 1877--1901. Curran Associates, Inc., 2020.

\bibitem[Chen et~al.(2015)Chen, Fang, Lin, Vedantam, Gupta, Doll{\'{a}}r, and Zitnick]{cocodata}
Xinlei Chen, Hao Fang, Tsung{-}Yi Lin, Ramakrishna Vedantam, Saurabh Gupta, Piotr Doll{\'{a}}r, and C.~Lawrence Zitnick.
\newblock Microsoft {COCO} captions: Data collection and evaluation server.
\newblock 2015.

\bibitem[Chowdhery et~al.(2022)Chowdhery, Narang, Devlin, Bosma, Mishra, Roberts, Barham, Chung, Sutton, Gehrmann, Schuh, Shi, Tsvyashchenko, Maynez, Rao, Barnes, Tay, Shazeer, Prabhakaran, Reif, Du, Hutchinson, Pope, Bradbury, Austin, Isard, Gur-Ari, Yin, Duke, Levskaya, Ghemawat, Dev, Michalewski, Garc{\'i}a, Misra, Robinson, Fedus, Zhou, Ippolito, Luan, Lim, Zoph, Spiridonov, Sepassi, Dohan, Agrawal, Omernick, Dai, Pillai, Pellat, Lewkowycz, Moreira, Child, Polozov, Lee, Zhou, Wang, Saeta, D{\'i}az, Firat, Catasta, Wei, Meier-Hellstern, Eck, Dean, Petrov, and Fiedel]{palm}
Aakanksha Chowdhery, Sharan Narang, Jacob Devlin, Maarten Bosma, Gaurav Mishra, Adam Roberts, Paul Barham, Hyung~Won Chung, Charles Sutton, Sebastian Gehrmann, Parker Schuh, Kensen Shi, Sasha Tsvyashchenko, Joshua Maynez, Abhishek~B Rao, Parker Barnes, Yi~Tay, Noam~M. Shazeer, Vinodkumar Prabhakaran, Emily Reif, Nan Du, Benton~C. Hutchinson, Reiner Pope, James Bradbury, Jacob Austin, Michael Isard, Guy Gur-Ari, Pengcheng Yin, Toju Duke, Anselm Levskaya, Sanjay Ghemawat, Sunipa Dev, Henryk Michalewski, Xavier Garc{\'i}a, Vedant Misra, Kevin Robinson, Liam Fedus, Denny Zhou, Daphne Ippolito, David Luan, Hyeontaek Lim, Barret Zoph, Alexander Spiridonov, Ryan Sepassi, David Dohan, Shivani Agrawal, Mark Omernick, Andrew~M. Dai, Thanumalayan~Sankaranarayana Pillai, Marie Pellat, Aitor Lewkowycz, Erica~Oliveira Moreira, Rewon Child, Oleksandr Polozov, Katherine Lee, Zongwei Zhou, Xuezhi Wang, Brennan Saeta, Mark D{\'i}az, Orhan Firat, Michele Catasta, Jason Wei, Kathleen~S. Meier-Hellstern, Douglas Eck, Jeff Dean,
  Slav Petrov, and Noah Fiedel.
\newblock {PaLM}: Scaling language modeling with pathways.
\newblock \emph{ArXiv}, abs/2204.02311, 2022.

\bibitem[Christiano et~al.(2017)Christiano, Leike, Brown, Martic, Legg, and Amodei]{christiano2017deep}
Paul~F Christiano, Jan Leike, Tom Brown, Miljan Martic, Shane Legg, and Dario Amodei.
\newblock Deep reinforcement learning from human preferences.
\newblock \emph{Advances in neural information processing systems}, 30, 2017.

\bibitem[Dai et~al.(2022)Dai, Dong, Hao, Sui, Chang, and Wei]{dai2022knowledge}
Damai Dai, Li~Dong, Yaru Hao, Zhifang Sui, Baobao Chang, and Furu Wei.
\newblock Knowledge neurons in pretrained transformers.
\newblock In \emph{Proceedings of the 60th Annual Meeting of the Association for Computational Linguistics (Volume 1: Long Papers)}, pages 8493--8502, Dublin, Ireland, May 2022. Association for Computational Linguistics.
\newblock \doi{10.18653/v1/2022.acl-long.581}.
\newblock URL \url{https://aclanthology.org/2022.acl-long.581}.

\bibitem[Deng et~al.(2009)Deng, Dong, Socher, Li, Li, and Fei-Fei]{imagenet}
Jia Deng, Wei Dong, Richard Socher, Li-Jia Li, Kai Li, and Li~Fei-Fei.
\newblock Imagenet: A large-scale hierarchical image database.
\newblock In \emph{2009 IEEE Conference on Computer Vision and Pattern Recognition}, pages 248--255, 2009.

\bibitem[Ding et~al.(2021)Ding, Yang, Hong, Zheng, Zhou, Yin, Lin, Zou, Shao, Yang, et~al.]{ding2021cogview}
Ming Ding, Zhuoyi Yang, Wenyi Hong, Wendi Zheng, Chang Zhou, Da~Yin, Junyang Lin, Xu~Zou, Zhou Shao, Hongxia Yang, et~al.
\newblock Cogview: Mastering text-to-image generation via transformers.
\newblock \emph{Advances in Neural Information Processing Systems}, 34:\penalty0 19822--19835, 2021.

\bibitem[Du et~al.(2021)Du, Huang, Dai, Tong, Lepikhin, Xu, Krikun, Zhou, Yu, Firat, Zoph, Fedus, Bosma, Zhou, Wang, Wang, Webster, Pellat, Robinson, Meier-Hellstern, Duke, Dixon, Zhang, Le, Wu, Chen, and Cui]{glam}
Nan Du, Yanping Huang, Andrew~M. Dai, Simon Tong, Dmitry Lepikhin, Yuanzhong Xu, Maxim Krikun, Yanqi Zhou, Adams~Wei Yu, Orhan Firat, Barret Zoph, Liam Fedus, Maarten Bosma, Zongwei Zhou, Tao Wang, Yu~Emma Wang, Kellie Webster, Marie Pellat, Kevin Robinson, Kathleen Meier-Hellstern, Toju Duke, Lucas Dixon, Kun Zhang, Quoc~V Le, Yonghui Wu, Zhifeng Chen, and Claire Cui.
\newblock Glam: Efficient scaling of language models with mixture-of-experts, 2021.

\bibitem[Gao et~al.(2021)Gao, Fisch, and Chen]{gao-etal-2021-making}
Tianyu Gao, Adam Fisch, and Danqi Chen.
\newblock Making pre-trained language models better few-shot learners.
\newblock In \emph{Proceedings of the 59th Annual Meeting of the Association for Computational Linguistics and the 11th International Joint Conference on Natural Language Processing (Volume 1: Long Papers)}, pages 3816--3830, Online, August 2021. Association for Computational Linguistics.
\newblock \doi{10.18653/v1/2021.acl-long.295}.
\newblock URL \url{https://aclanthology.org/2021.acl-long.295}.

\bibitem[Gu et~al.(2022)Gu, Chen, Bao, Wen, Zhang, Chen, Yuan, and Guo]{gu2022vector}
Shuyang Gu, Dong Chen, Jianmin Bao, Fang Wen, Bo~Zhang, Dongdong Chen, Lu~Yuan, and Baining Guo.
\newblock Vector quantized diffusion model for text-to-image synthesis.
\newblock In \emph{Proceedings of the IEEE/CVF Conference on Computer Vision and Pattern Recognition}, pages 10696--10706, 2022.

\bibitem[Hancock et~al.(2019)Hancock, Bordes, Mazare, and Weston]{hancock-etal-2019-learning}
Braden Hancock, Antoine Bordes, Pierre-Emmanuel Mazare, and Jason Weston.
\newblock Learning from dialogue after deployment: Feed yourself, chatbot!
\newblock In \emph{Proceedings of the 57th Annual Meeting of the Association for Computational Linguistics}, pages 3667--3684, Florence, Italy, July 2019. Association for Computational Linguistics.
\newblock \doi{10.18653/v1/P19-1358}.
\newblock URL \url{https://aclanthology.org/P19-1358}.

\bibitem[Haviv et~al.(2021)Haviv, Berant, and Globerson]{haviv-etal-2021-bertese}
Adi Haviv, Jonathan Berant, and Amir Globerson.
\newblock {BERT}ese: Learning to speak to {BERT}.
\newblock In \emph{Proceedings of the 16th Conference of the European Chapter of the Association for Computational Linguistics: Main Volume}, pages 3618--3623, Online, April 2021. Association for Computational Linguistics.
\newblock \doi{10.18653/v1/2021.eacl-main.316}.
\newblock URL \url{https://aclanthology.org/2021.eacl-main.316}.

\bibitem[Ibarz et~al.(2018)Ibarz, Leike, Pohlen, Irving, Legg, and Amodei]{ibarz2018reward}
Borja Ibarz, Jan Leike, Tobias Pohlen, Geoffrey Irving, Shane Legg, and Dario Amodei.
\newblock Reward learning from human preferences and demonstrations in atari.
\newblock In S.~Bengio, H.~Wallach, H.~Larochelle, K.~Grauman, N.~Cesa-Bianchi, and R.~Garnett, editors, \emph{Advances in Neural Information Processing Systems}, volume~31. Curran Associates, Inc., 2018.
\newblock URL \url{https://proceedings.neurips.cc/paper/2018/file/8cbe9ce23f42628c98f80fa0fac8b19a-Paper.pdf}.

\bibitem[Jaques et~al.(2019)Jaques, Ghandeharioun, Shen, Ferguson, Lapedriza, Jones, Gu, and Picard]{jaques2019way}
Natasha Jaques, Asma Ghandeharioun, Judy~Hanwen Shen, Craig Ferguson, Agata Lapedriza, Noah Jones, Shixiang Gu, and Rosalind Picard.
\newblock Way off-policy batch deep reinforcement learning of implicit human preferences in dialog.
\newblock \emph{arXiv preprint arXiv:1907.00456}, 2019.

\bibitem[Jiang et~al.(2020)Jiang, Xu, Araki, and Neubig]{jiang2020how}
Zhengbao Jiang, Frank~F. Xu, Jun Araki, and Graham Neubig.
\newblock {How Can We Know What Language Models Know?}
\newblock \emph{Transactions of the Association for Computational Linguistics}, 8:\penalty0 423--438, 07 2020.
\newblock ISSN 2307-387X.
\newblock \doi{10.1162/tacl_a_00324}.
\newblock URL \url{https://doi.org/10.1162/tacl\_a\_00324}.

\bibitem[Klie et~al.(2020)Klie, Eckart~de Castilho, and Gurevych]{klie-etal-2020-zero}
Jan-Christoph Klie, Richard Eckart~de Castilho, and Iryna Gurevych.
\newblock {F}rom {Z}ero to {H}ero: {H}uman-{I}n-{T}he-{L}oop {E}ntity {L}inking in {L}ow {R}esource {D}omains.
\newblock In \emph{Proceedings of the 58th Annual Meeting of the Association for Computational Linguistics}, pages 6982--6993, Online, July 2020. Association for Computational Linguistics.
\newblock \doi{10.18653/v1/2020.acl-main.624}.
\newblock URL \url{https://aclanthology.org/2020.acl-main.624}.

\bibitem[Li and Liang(2021)]{li-liang-2021-prefix}
Xiang~Lisa Li and Percy Liang.
\newblock Prefix-tuning: Optimizing continuous prompts for generation.
\newblock In \emph{Proceedings of the 59th Annual Meeting of the Association for Computational Linguistics and the 11th International Joint Conference on Natural Language Processing (Volume 1: Long Papers)}, pages 4582--4597, Online, August 2021. Association for Computational Linguistics.
\newblock \doi{10.18653/v1/2021.acl-long.353}.
\newblock URL \url{https://aclanthology.org/2021.acl-long.353}.

\bibitem[Liu and Chilton(2021)]{promptguideline}
Vivian Liu and Lydia~B. Chilton.
\newblock Design guidelines for prompt engineering text-to-image generative models, 2021.

\bibitem[Liu et~al.(2019)Liu, Ott, Goyal, Du, Joshi, Chen, Levy, Lewis, Zettlemoyer, and Stoyanov]{roberta}
Yinhan Liu, Myle Ott, Naman Goyal, Jingfei Du, Mandar Joshi, Danqi Chen, Omer Levy, Mike Lewis, Luke Zettlemoyer, and Veselin Stoyanov.
\newblock {RoBERTa}: A robustly optimized bert pretraining approach.
\newblock \emph{arXiv preprint arXiv:1907.11692}, 2019.

\bibitem[Lu et~al.(2022)Lu, Zhou, Bao, Chen, Li, and Zhu]{dpmsolver}
Cheng Lu, Yuhao Zhou, Fan Bao, Jianfei Chen, Chongxuan Li, and Jun Zhu.
\newblock Dpm-solver: A fast ode solver for diffusion probabilistic model sampling in around 10 steps.
\newblock \emph{arXiv preprint arXiv:2206.00927}, 2022.

\bibitem[Murray et~al.(2012)Murray, Marchesotti, and Perronnin]{murray2012ava}
Naila Murray, Luca Marchesotti, and Florent Perronnin.
\newblock Ava: A large-scale database for aesthetic visual analysis.
\newblock In \emph{2012 IEEE conference on computer vision and pattern recognition}, pages 2408--2415. IEEE, 2012.

\bibitem[Oppenlaender(2022)]{promptmodifier}
Jonas Oppenlaender.
\newblock A taxonomy of prompt modifiers for text-to-image generation, 2022.

\bibitem[Ouyang et~al.(2022)Ouyang, Wu, Jiang, Almeida, Wainwright, Mishkin, Zhang, Agarwal, Slama, Ray, et~al.]{instructgpt}
Long Ouyang, Jeff Wu, Xu~Jiang, Diogo Almeida, Carroll~L Wainwright, Pamela Mishkin, Chong Zhang, Sandhini Agarwal, Katarina Slama, Alex Ray, et~al.
\newblock Training language models to follow instructions with human feedback.
\newblock \emph{arXiv preprint arXiv:2203.02155}, 2022.

\bibitem[Parsons(2022)]{guidebook}
Guy Parsons.
\newblock The dall·e 2 prompt book, 2022.

\bibitem[Petroni et~al.(2019)Petroni, Rockt{\"a}schel, Riedel, Lewis, Bakhtin, Wu, and Miller]{petroni2019language}
Fabio Petroni, Tim Rockt{\"a}schel, Sebastian Riedel, Patrick Lewis, Anton Bakhtin, Yuxiang Wu, and Alexander Miller.
\newblock Language models as knowledge bases?
\newblock In \emph{Proceedings of the 2019 Conference on Empirical Methods in Natural Language Processing and the 9th International Joint Conference on Natural Language Processing (EMNLP-IJCNLP)}, pages 2463--2473, Hong Kong, China, November 2019. Association for Computational Linguistics.
\newblock \doi{10.18653/v1/D19-1250}.
\newblock URL \url{https://aclanthology.org/D19-1250}.

\bibitem[Radford et~al.(2019)Radford, Wu, Child, Luan, Amodei, and Sutskever]{gpt2}
Alec Radford, Jeff Wu, Rewon Child, David Luan, Dario Amodei, and Ilya Sutskever.
\newblock Language models are unsupervised multitask learners.
\newblock 2019.

\bibitem[Radford et~al.(2021)Radford, Kim, Hallacy, Ramesh, Goh, Agarwal, Sastry, Askell, Mishkin, Clark, Krueger, and Sutskever]{clip}
Alec Radford, Jong~Wook Kim, Chris Hallacy, Aditya Ramesh, Gabriel Goh, Sandhini Agarwal, Girish Sastry, Amanda Askell, Pamela Mishkin, Jack Clark, Gretchen Krueger, and Ilya Sutskever.
\newblock Learning transferable visual models from natural language supervision.
\newblock In Marina Meila and Tong Zhang, editors, \emph{Proceedings of the 38th International Conference on Machine Learning}, volume 139 of \emph{Proceedings of Machine Learning Research}, pages 8748--8763. PMLR, 18--24 Jul 2021.
\newblock URL \url{https://proceedings.mlr.press/v139/radford21a.html}.

\bibitem[Ramesh et~al.(2021{\natexlab{a}})Ramesh, Pavlov, Goh, Gray, Voss, Radford, Chen, and Sutskever]{dalle}
Aditya Ramesh, Mikhail Pavlov, Gabriel Goh, Scott Gray, Chelsea Voss, Alec Radford, Mark Chen, and Ilya Sutskever.
\newblock Zero-shot text-to-image generation.
\newblock \emph{arXiv}, abs/2102.12092, 2021{\natexlab{a}}.

\bibitem[Ramesh et~al.(2021{\natexlab{b}})Ramesh, Pavlov, Goh, Gray, Voss, Radford, Chen, and Sutskever]{ramesh2021zero}
Aditya Ramesh, Mikhail Pavlov, Gabriel Goh, Scott Gray, Chelsea Voss, Alec Radford, Mark Chen, and Ilya Sutskever.
\newblock Zero-shot text-to-image generation.
\newblock In \emph{International Conference on Machine Learning}, pages 8821--8831. PMLR, 2021{\natexlab{b}}.

\bibitem[Ramesh et~al.(2022)Ramesh, Dhariwal, Nichol, Chu, and Chen]{dalle2}
Aditya Ramesh, Prafulla Dhariwal, Alex Nichol, Casey Chu, and Mark Chen.
\newblock Hierarchical text-conditional image generation with clip latents, 2022.

\bibitem[Reed et~al.(2016{\natexlab{a}})Reed, Akata, Yan, Logeswaran, Schiele, and Lee]{reed2016generative}
Scott Reed, Zeynep Akata, Xinchen Yan, Lajanugen Logeswaran, Bernt Schiele, and Honglak Lee.
\newblock Generative adversarial text to image synthesis.
\newblock In \emph{International conference on machine learning}, pages 1060--1069. PMLR, 2016{\natexlab{a}}.

\bibitem[Reed et~al.(2016{\natexlab{b}})Reed, Akata, Mohan, Tenka, Schiele, and Lee]{reed2016learning}
Scott~E Reed, Zeynep Akata, Santosh Mohan, Samuel Tenka, Bernt Schiele, and Honglak Lee.
\newblock Learning what and where to draw.
\newblock \emph{Advances in neural information processing systems}, 29, 2016{\natexlab{b}}.

\bibitem[Reynolds and McDonell(2021)]{promptprogramming}
Laria Reynolds and Kyle McDonell.
\newblock Prompt programming for large language models: Beyond the few-shot paradigm, 2021.

\bibitem[Rombach et~al.(2022)Rombach, Blattmann, Lorenz, Esser, and Ommer]{sdiffusion}
Robin Rombach, Andreas Blattmann, Dominik Lorenz, Patrick Esser, and Bj\"orn Ommer.
\newblock High-resolution image synthesis with latent diffusion models.
\newblock In \emph{Proceedings of the IEEE/CVF Conference on Computer Vision and Pattern Recognition (CVPR)}, pages 10684--10695, June 2022.

\bibitem[Saharia et~al.(2022)Saharia, Chan, Saxena, Li, Whang, Denton, Ghasemipour, Ayan, Mahdavi, Lopes, Salimans, Ho, Fleet, and Norouzi]{imagegen}
Chitwan Saharia, William Chan, Saurabh Saxena, Lala Li, Jay Whang, Emily Denton, Seyed Kamyar~Seyed Ghasemipour, Burcu~Karagol Ayan, S.~Sara Mahdavi, Rapha~Gontijo Lopes, Tim Salimans, Jonathan Ho, David~J Fleet, and Mohammad Norouzi.
\newblock Photorealistic text-to-image diffusion models with deep language understanding, 2022.

\bibitem[Schulman et~al.(2017)Schulman, Wolski, Dhariwal, Radford, and Klimov]{ppo}
John Schulman, Filip Wolski, Prafulla Dhariwal, Alec Radford, and Oleg Klimov.
\newblock Proximal policy optimization algorithms.
\newblock \emph{ArXiv}, abs/1707.06347, 2017.

\bibitem[Shin et~al.(2021)Shin, Lin, Thomson, Chen, Roy, Platanios, Pauls, Klein, Eisner, and Van~Durme]{shin-etal-2021-constrained}
Richard Shin, Christopher Lin, Sam Thomson, Charles Chen, Subhro Roy, Emmanouil~Antonios Platanios, Adam Pauls, Dan Klein, Jason Eisner, and Benjamin Van~Durme.
\newblock Constrained language models yield few-shot semantic parsers.
\newblock In \emph{Proceedings of the 2021 Conference on Empirical Methods in Natural Language Processing}, pages 7699--7715, Online and Punta Cana, Dominican Republic, November 2021. Association for Computational Linguistics.
\newblock \doi{10.18653/v1/2021.emnlp-main.608}.
\newblock URL \url{https://aclanthology.org/2021.emnlp-main.608}.

\bibitem[Shuster et~al.(2020)Shuster, Urbanek, Dinan, Szlam, and Weston]{shuster2020deploying}
Kurt Shuster, Jack Urbanek, Emily Dinan, Arthur Szlam, and Jason Weston.
\newblock Deploying lifelong open-domain dialogue learning.
\newblock \emph{arXiv preprint arXiv:2008.08076}, 2020.

\bibitem[Smith et~al.(2022)Smith, Patwary, Norick, LeGresley, Rajbhandari, Casper, Liu, Prabhumoye, Zerveas, Korthikanti, Zhang, Child, Aminabadi, Bernauer, Song, Shoeybi, He, Houston, Tiwary, and Catanzaro]{mtnlg}
Shaden Smith, Mostofa Patwary, Brandon Norick, Patrick LeGresley, Samyam Rajbhandari, Jared Casper, Zhun Liu, Shrimai Prabhumoye, George Zerveas, Vijay Korthikanti, Elton Zhang, Rewon Child, Reza~Yazdani Aminabadi, Julie Bernauer, Xia Song, Mohammad Shoeybi, Yuxiong He, Michael Houston, Saurabh Tiwary, and Bryan Catanzaro.
\newblock Using {DeepSpeed} and {Megatron} to train {Megatron-Turing NLG 530B}, a large-scale generative language model, 2022.

\bibitem[Stiennon et~al.(2020)Stiennon, Ouyang, Wu, Ziegler, Lowe, Voss, Radford, Amodei, and Christiano]{learningtosummarize}
Nisan Stiennon, Long Ouyang, Jeffrey Wu, Daniel~M. Ziegler, Ryan Lowe, Chelsea Voss, Alec Radford, Dario Amodei, and Paul~F. Christiano.
\newblock Learning to summarize with human feedback.
\newblock In \emph{Advances in Neural Information Processing Systems 33: Annual Conference on Neural Information Processing Systems 2020, NeurIPS 2020, December 6-12, 2020, virtual}, 2020.

\bibitem[Tao et~al.(2022)Tao, Tang, Wu, Jing, Bao, and Xu]{tao2022df}
Ming Tao, Hao Tang, Fei Wu, Xiao-Yuan Jing, Bing-Kun Bao, and Changsheng Xu.
\newblock Df-gan: A simple and effective baseline for text-to-image synthesis.
\newblock In \emph{Proceedings of the IEEE/CVF Conference on Computer Vision and Pattern Recognition}, pages 16515--16525, 2022.

\bibitem[Tsimpoukelli et~al.(2021)Tsimpoukelli, Menick, Cabi, Eslami, Vinyals, and Hill]{tsimpoukelli2021multimodal}
Maria Tsimpoukelli, Jacob~L Menick, Serkan Cabi, SM~Eslami, Oriol Vinyals, and Felix Hill.
\newblock Multimodal few-shot learning with frozen language models.
\newblock \emph{Advances in Neural Information Processing Systems}, 34:\penalty0 200--212, 2021.

\bibitem[Vaswani et~al.(2017)Vaswani, Shazeer, Parmar, Uszkoreit, Jones, Gomez, Kaiser, and Polosukhin]{transformer}
Ashish Vaswani, Noam Shazeer, Niki Parmar, Jakob Uszkoreit, Llion Jones, Aidan~N Gomez, {\L}ukasz Kaiser, and Illia Polosukhin.
\newblock Attention is all you need.
\newblock In \emph{Advances in Neural Information Processing Systems}, pages 5998--6008. Curran Associates, Inc., 2017.
\newblock URL \url{http://papers.nips.cc/paper/7181-attention-is-all-you-need.pdf}.

\bibitem[Vijayakumar et~al.(2016)Vijayakumar, Cogswell, Selvaraju, Sun, Lee, Crandall, and Batra]{dbs}
Ashwin~K. Vijayakumar, Michael Cogswell, Ramprasaath~R. Selvaraju, Qing Sun, Stefan Lee, David~J. Crandall, and Dhruv Batra.
\newblock Diverse beam search: Decoding diverse solutions from neural sequence models.
\newblock \emph{ArXiv}, abs/1610.02424, 2016.

\bibitem[Wang et~al.(2022)Wang, Montoya, Munechika, Yang, Hoover, and Chau]{diffusiondb}
Zijie~J. Wang, Evan Montoya, David Munechika, Haoyang Yang, Benjamin Hoover, and Duen~Horng Chau.
\newblock {{DiffusionDB}}: {{A}} large-scale prompt gallery dataset for text-to-image generative models.
\newblock \emph{arXiv:2210.14896 [cs]}, 2022.
\newblock URL \url{https://arxiv.org/abs/2210.14896}.

\bibitem[Wei et~al.(2022)Wei, Wang, Schuurmans, Bosma, brian ichter, Xia, Chi, Le, and Zhou]{wei2022chain}
Jason Wei, Xuezhi Wang, Dale Schuurmans, Maarten Bosma, brian ichter, Fei Xia, Ed~H. Chi, Quoc~V Le, and Denny Zhou.
\newblock Chain of thought prompting elicits reasoning in large language models.
\newblock In Alice~H. Oh, Alekh Agarwal, Danielle Belgrave, and Kyunghyun Cho, editors, \emph{Advances in Neural Information Processing Systems}, 2022.
\newblock URL \url{https://openreview.net/forum?id=_VjQlMeSB_J}.

\bibitem[Xu et~al.(2022)Xu, Ung, Komeili, Arora, Boureau, and Weston]{xu2022learning}
Jing Xu, Megan Ung, Mojtaba Komeili, Kushal Arora, Y-Lan Boureau, and Jason Weston.
\newblock Learning new skills after deployment: Improving open-domain internet-driven dialogue with human feedback.
\newblock \emph{arXiv preprint arXiv:2208.03270}, 2022.

\bibitem[Yao et~al.(2019)Yao, Su, Sun, and Yih]{yao-etal-2019-model}
Ziyu Yao, Yu~Su, Huan Sun, and Wen-tau Yih.
\newblock Model-based interactive semantic parsing: A unified framework and a text-to-{SQL} case study.
\newblock In \emph{Proceedings of the 2019 Conference on Empirical Methods in Natural Language Processing and the 9th International Joint Conference on Natural Language Processing (EMNLP-IJCNLP)}, pages 5447--5458, Hong Kong, China, November 2019. Association for Computational Linguistics.
\newblock \doi{10.18653/v1/D19-1547}.
\newblock URL \url{https://aclanthology.org/D19-1547}.

\bibitem[Zhou et~al.(2022{\natexlab{a}})Zhou, Yang, Loy, and Liu]{zhou2022learning}
Kaiyang Zhou, Jingkang Yang, Chen~Change Loy, and Ziwei Liu.
\newblock Learning to prompt for vision-language models.
\newblock \emph{International Journal of Computer Vision}, 130\penalty0 (9):\penalty0 2337--2348, 2022{\natexlab{a}}.

\bibitem[Zhou et~al.(2022{\natexlab{b}})Zhou, Muresanu, Han, Paster, Pitis, Chan, and Ba]{lmengineer}
Yongchao Zhou, Andrei~Ioan Muresanu, Ziwen Han, Keiran Paster, Silviu Pitis, Harris Chan, and Jimmy Ba.
\newblock Large language models are human-level prompt engineers, 2022{\natexlab{b}}.

\bibitem[Ziegler et~al.(2019)Ziegler, Stiennon, Wu, Brown, Radford, Amodei, Christiano, and Irving]{ziegler2019fine}
Daniel~M Ziegler, Nisan Stiennon, Jeffrey Wu, Tom~B Brown, Alec Radford, Dario Amodei, Paul Christiano, and Geoffrey Irving.
\newblock Fine-tuning language models from human preferences.
\newblock \emph{arXiv preprint arXiv:1909.08593}, 2019.

\end{thebibliography}
